\documentclass{article}

\usepackage{microtype}
\usepackage{graphicx}
\usepackage{subfigure}
\usepackage{booktabs} 
\usepackage{array}
\usepackage{multirow}
\usepackage{bm}

\usepackage{hyperref}

\usepackage[accepted]{icml2024}

\usepackage{amsmath}
\usepackage{amssymb}
\usepackage{mathtools}
\usepackage{amsthm}

\usepackage[capitalize,noabbrev]{cleveref}

\theoremstyle{plain}

\theoremstyle{definition}

\theoremstyle{remark}

\usepackage[textsize=tiny]{todonotes}

\icmltitlerunning{Copyright Traps for Large Language Models}

\begin{document}

\twocolumn[
\icmltitle{Copyright Traps for Large Language Models}



\icmlsetsymbol{equal}{*}

\begin{icmlauthorlist}
\icmlauthor{Matthieu Meeus}{equal,imperial}
\icmlauthor{Igor Shilov}{equal,imperial}
\icmlauthor{Manuel Faysse}{central}
\icmlauthor{Yves-Alexandre de Montjoye}{imperial}
\end{icmlauthorlist}

\icmlaffiliation{imperial}{Department of Computing, Imperial College London, United Kingdom}
\icmlaffiliation{central}{MICS, CentraleSupélec, Université Paris-Saclay, Paris, France}

\icmlcorrespondingauthor{Yves-Alexandre de Montjoye}{deMontjoye@imperial.ac.uk}

\icmlkeywords{Machine Learning, ICML}

\vskip 0.3in
]



\printAffiliationsAndNotice{\icmlEqualContribution} 

\begin{abstract}

Questions of fair use of copyright-protected content to train Large Language Models (LLMs) are being actively debated. Document-level inference has been proposed as a new task: inferring from black-box access to the trained model whether a piece of content has been seen during training. SOTA methods however rely on naturally occurring memorization of (part of) the content. While very effective against models that memorize significantly, we hypothesize--and later confirm--that they will not work against models that do not naturally memorize, e.g. medium-size 1B models. We here propose to use copyright traps, the inclusion of fictitious entries in original content, to detect the use of copyrighted materials in LLMs with a focus on models where memorization does not naturally occur. We carefully design a randomized controlled experimental setup, inserting traps into original content (books) and train a 1.3B LLM from scratch. We first validate that the use of content in our target model would be undetectable using existing methods. We then show, contrary to intuition, that even medium-length trap sentences repeated a significant number of times (100) are not detectable using existing methods. However, we show that longer sequences repeated a large number of times can be reliably detected (AUC=0.75) and used as copyright traps. 
Beyond copyright applications, our findings contribute to the study of LLM memorization: the randomized controlled setup enables us to draw causal relationships between memorization and certain sequence properties such as repetition in model training data and perplexity.

\end{abstract}

\section{Introduction}

With the growing adoption of ever-improving Large Language Models (LLMs), concerns are being raised when it comes to the use of copyright protected content for training. 
Numerous content creators have indeed filed lawsuits against technology companies, claiming copyright infringement for utilizing books~\cite{authorsguild,silvermanmeta}, songs~\cite{anthropic} or news articles~\cite{nytimes} for LLM development. While it is still unclear whether copyright or \emph{fair use} applies in this context~\cite{samuelson2023generative}, model developers continue releasing new LLMs but are increasingly reluctant to disclose details on the training dataset~\cite{gpt4techreport,touvron2023llama2,jiang2024mixtral} - partially due to these lawsuits. 

Methods have recently been developed to detect whether a specific piece of content has been seen by an LLM during training: document-level membership inference. Both~\cite{meeus2023did} and~\cite{shi2023detecting} show their methods to be fairly successful against very large LLMs (up to 66B parameters), with a ROC AUC of 0.86 for OpenLLaMA~\cite{openlm2023openllama} and 0.88 for GPT-3~\cite{brown2020language}.

\begin{figure}[t]
\centering
\includegraphics[width=0.6\linewidth]{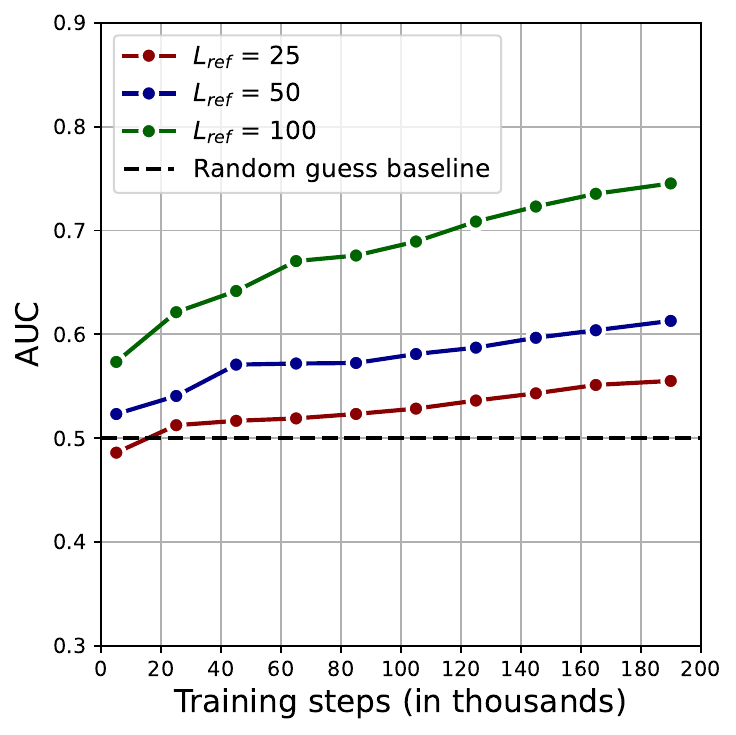} 
    \caption{\textbf{Memorization throughout training.} The \textit{Ratio} MIA performance (AUC) for synthetically generated trap sequences (of varying sequence length), repeated 1,000 times in a book, evaluated on intermediate checkpoints of the target LLM.}
\label{fig:across_training}
\end{figure}

Historically, original content creators have implemented so-called \emph{copyright traps} to detect copyright infringement of their work. Examples of such traps range from a fictitious street name or town on a map to the inclusion of fabricated names in a dictionary~\cite{Alford_2005a}. In this case, the direct inclusion of these entities in other work would render a breach of copyright self-evident, while it becomes less trivial when data is aggregated, e.g. when used in machine learning models. 

We here investigate, for the first time, the use of copyright traps for document-level membership inference against LLMs. We propose the injection of purposefully designed text (\emph{trap sequences}) into a piece of content, to either further improve the performance of document-level membership inference or enable it in the first place for models less prone to memorization. 

We here focus on the latter, as recently proposed methods are already successful for larger models~\cite{meeus2023did,shi2023detecting}, and as there is a growing trend towards smaller language models~\cite{zhang2024tinyllama,Javaheripi_2023}. 

Specifically, we inject our traps into the training set of CroissantLLM~\cite{faysse2024croissantllm}, a 1.3B parameter LLM, trained from scratch on 3 trillion tokens by the team we partnered with. Being (fairly) small and trained on significantly more data than considered in prior work on LLM memorization~\cite{carlini2022quantifying}, we hypothesized that the model would not naturally memorize sufficiently for a document-level membership inference to succeed. Applying the two state-of-the-art methods~\cite{meeus2023did,shi2023detecting}, we find them to perform barely better than a random guess baseline, confirming our hypothesis and rendering these methods uninformative for authors. 

We hence investigate the use of document-specific copyright traps to enable membership inference. We apply Membership Inference Attacks (MIAs) from the literature~\cite{yeom2018privacy,carlini2021extracting,shi2023detecting} to infer whether a given trap sequence, and thus document, has been seen by a model or not.

First, we consider synthetically generated trap sequences and study the impact of number of repetitions, sequence length, and perplexity  on the post training detectability of the trap. Contrary to popular beliefs, notably from the training data extraction literature~\cite{carlini2021extracting,nasr2023scalable}, we show that short and medium-length synthetic sequences repeated a significant number of times (100) do not help the membership inference, independently of the detection method used. We further confirm this also holds for artificially duplicated existing sentences. 

We, however, do find that the MIA AUC increases with sequence length and number of repetitions and that sequences of 100 tokens repeated 1,000 times are detectable with an AUC of 0.748. This provides the first evidence that copyright traps can be inserted in real-world LLMs to detect the use of training content otherwise undetectable.


We also show that sequences with high-perplexity (according to a reference model) are more likely to be detectable. The general intuition is that 'outliers' might more easily be memorized and be more vulnerable against MIAs~\cite{feldman2020does}. We are the first to test this out for LLMs in a clean setup, and show that when memorization happens (for long sequences repeated 1,000 times), the MIA AUC improves from approximately 0.65 for low perplexity to 0.8 for high perplexity. We also show the relationship between perplexity and detectability to be a potential confounding factor in prior post-hoc studies of LLM memorization, by studying the perplexity of duplicate sequences in the large text dataset The Pile~\cite{pile}.

Our results provide the first evidence that target-model independent copyright traps can be added to content to enable document-level membership inference, even in LLMs that would not 'naturally' memorize sufficiently to infer membership. 

While injecting traps might be not be equally trivial across document types while maintaining readability, they can be embedded across a large corpus (e.g. news articles). They can also be hidden online and not trivial to remove, especially given automated scraping and the costs associated with fine-grained deduplication for LLM training data.


\section{Related work}

\subsection{Document-level MIAs for LLMs}

With model developers becoming more reluctant to disclose details on their training sources~\cite{bommasani2023foundation}, partially due to copyright concerns raised by content creators~\cite{atlantic, silvermanmeta}, research has emerged recently aiming to infer whether a model of interest has been trained on a particular piece of text.~\cite{meeus2023did} has proposed a document-level MIA -leveraging the collection of member and non-member documents and a meta-classifier- and demonstrated its effectiveness in inferring membership for documents (books, papers) used to train OpenLLaMA~\cite{openlm2023openllama}.~\cite{shi2023detecting} uses a similar membership dataset collection strategy and successfully applied their novel sequence-level MIA to the same document-level membership inference task on GPT-3~\cite{brown2020language}.

Contrary to our work, both techniques rely on naturally occurring memorization. We instead propose to modify the document in a way that enables detectability even in models that do not naturally memorize. 

\subsection{Privacy attacks in a controlled setup}

Membership Inference Attacks (MIAs) have long been used in the privacy literature. They were originally introduced to infer the contribution of an individual sample in data aggregates~\cite{homer2008resolving} and have been expanded to machine learning (ML) models and other aggregation techniques~\cite{shokri2017membership,pyrgelis2017knock}. 

MIAs against ML models have been implemented under a wide range of assumptions made for the attacker, ranging from white-box access to the target model~\cite{nasr2018comprehensive,sablayrolles2019white,cretu2023re} to black-box access to the model confidence vector~\cite{shokri2017membership} to access to the predicted labels only~\cite{choquette2021label}.  

MIAs often leverage the shadow modeling setup, where multiple models are trained on datasets either including or excluding the record of interest. This allows for a controlled experiment setup, eliminating potential bias in the data. The decision boundary for membership can then either be inferred by a binary meta-classifier~\cite{shokri2017membership,meeus2023achilles} or through metrics computed on the model output~\cite{yeom2018privacy,carlini2022membership}.

Beyond MIAs, prior work have used injection techniques to study training data extraction attacks against small scale language models~\cite{henderson2018ethical, thakkar2020understanding, thomas2020investigating}. Notably,~\cite{carlini2019secret} generates hand-crafted canaries containing ``secret'' information (e.g. "my credit card number is " followed by a set of 9 digits) and proposes an \textit{exposure} metric to quantify the memorization. 

\subsection{Measuring naturally occurring LLM memorization}

MIAs have also been used to study naturally occurring memorization in LLMs at the sequence level. Some methods leverage shadow models~\cite{song2019auditing,hisamoto2020membership,carlini2022membership}, but the computational cost to train modern LLMs~\cite{radford2019language,touvron2023llama} has rendered them impractical. Novel MIAs thus use the model loss~\cite{yeom2018privacy}, leverage the access to one reference model~\cite{mireshghallah2022quantifying}, assume access to the model weights~\cite{li2023mope}, or generate \emph{neighboring samples} and predict membership based on the model loss of these samples~\cite{mattern2023membership}.~\cite{kandpal2022deduplicating}, for instance, uses some of these methods to demonstrate that data duplication is a major contributing factor to training data memorization. 

Beyond MIAs, the problem of training data extraction has been studied extensively in recent years. While earlier research focused on the qualitative demonstration that extraction is possible~\cite{carlini2019secret, carlini2021extracting}, more recent work has looked increasingly into quantitatively measuring the extent to which models memorize and factors contributing to higher memorization~\cite{carlini2022quantifying, kandpal2022deduplicating}. 

All of the studies focusing on LLM memorization furthermore rely on naturally occurring memorization. While the computational cost to train LLMs might inhibit a fully randomized and controlled setup, the lack of randomization means that confounding factors might, possibly strongly, impact the results. For instance, sequences repeated more often might be the footer added by a publisher to every book while a sequence repeated only a few times might come from a book which appears multiple times in the dataset. In this case, the relationship between duplication and memorization will likely be strongly impacted by sequence type and context, introducing potential measurement bias in the results. 

We here, for the first time, uniquely train an LLM from scratch while randomly injecting, in particular synthetic, trap sequences. While not our primary goal, we expect our release of trap sequences and the target model to provide a fully randomized controlled setup to understand LLM memorization - beyond the document-level inference task considered here.

\label{sec:related_work}

\section{Preliminary}
\subsection{Language modeling}
 As target model, we consider an autoregressive large language model \textit{LM}, i.e. trained for next-token prediction. Model parameters $\theta$ are determined by minimizing the cross-entropy loss for the predicted probability distribution for the next token given preceding tokens, for the entire training dataset $\mathcal{D}_{\text{train}}$. 

We denote the corresponding tokenizer as $T$. A sequence of textual characters $X$ can then be encoded using $T$ to a sequence of $L$ tokens, $T(X) = \{t_1,\ldots,t_L\}$. 

The model loss for this sequence is computed as follows: 

\begin{equation}
\begin{aligned}
\label{eq:loss}
\mathcal{L}_{\textit{LM}}(X) & = -\frac{1}{L}\sum_{i=1}^{L} \log\left( \textit{LM}_{\theta}(t_i | t_1 \ldots, t_{i-1})\right) \\
& = -\frac{1}{L}\sum_{i=1}^{L} \log\left( \textit{LM}_{\theta}(t_i)\right)
\end{aligned} 
\end{equation}

Here $\textit{LM}_{\theta}(t_i)$ represents the predicted probability for token $t_i$ returned by model \textit{LM} with parameters $\theta$ and context $(t_1 \ldots, t_{i-1})$. The \emph{perplexity} of a sequence $X$ is computed as the exponent of the loss, or 
$\mathcal{P}_{\textit{LM}}(X) = \exp\left(\mathcal{L}_{\textit{LM}}(X)\right)$. 

\subsection{Threat model}
\label{sec:threat_model}

We consider as attacker an original content creator who is in possession of an original document $D$ (or set of documents) that might be used to train an LLM. 

We further assume the attacker to have black-box access to a reference language model $\textit{LM}_{\text{ref}}$ with tokenizer $T_{\text{ref}}$, which is reasonable to assume with many LLMs publicly available~\cite{touvron2023llama2,scao2022bloom,jiang2024mixtral}. This also includes the ability to generate synthetic sequences using $\textit{LM}_{\text{ref}}$ as explained in Sec.~\ref{sec:m_generation}. 

In our setup, the attacker injects a sequence of textual characters -the trap sequence $M_D$, which is unique to this document $D$- to create the modified document $D'$, where: 

\begin{enumerate}
    \item The length of $M_D$ is defined by the tokenizer of the reference model and denoted as $L_{\text{ref}}(M_D) = |T_{\text{ref}}(M_D)|$.
    \item The perplexity of $M_D$ is computed by the reference model and denoted as $\mathcal{P}_{\textit{LM}_{\text{ref}}}(M_D)$.
    \item Modified document $D'$ is obtained by randomly injecting the textual characters $M_D$ an $n_{\text{rep}}$ number of times into the original document $D$.
\end{enumerate}

We assume the modified document $D'$ made available for a wider audience, including potential LLM developers. 

The target model for the attacker is the language model \textit{LM} that has been pretrained on dataset $\mathcal{D}_{\text{train}}$. We also assume the attacker to have black-box access to \textit{LM}. The attacker's goal is now to infer document-level membership, i.e. whether their modified document $D'$ has been used to train \textit{LM} (in other words, if $D' \in \mathcal{D}_{\text{train}}$ or not). Importantly for our experimental results, as the trap sequence $M_D$ is unique to the document $D$, we perform a sequence-level MIA for the trap sequence $M_D$ as a lower bound approximation for the document-level membership inference. We here use \textit{detectability} to refer to the ability to detect that a trap has been seen by language model \textit{LM} during training.
\label{sec:preliminary}


\section{Experiment Design}
\subsection{Trap sequence generation}
\label{sec:m_generation}

We construct trap sequences controlling for:

\textbf{\textit{1. Sequence length}} in tokens using the tokenizer of the reference model, or $L_{\text{ref}}(M_D)$. We consider $L_{\text{ref}}(M_D) = \{25, 50, 100\}$ tokens.

\textbf{\textit{2. Perplexity}} according to the reference model. We define 10 \emph{perplexity buckets} $b_i$, such that $\forall{\mathcal{P}_{\textit{LM}_{\text{ref}}}(M_D) \in b_i}$: $1 + (i - 1) \cdot 10 \leq \mathcal{P}_{\textit{LM}_{\text{ref}}}(M_D) < 1 + i \cdot 10$ for $i= 1 \ldots 10$. 

We hypothesize that both properties have an impact on memorization. For the sequence length, prior work~\cite{carlini2022quantifying} showed in a post-hoc analysis that longer sequences are consistently more extractable. For perplexity, we base this on the intuition that perplexity captures the model's surprise, and the higher-perplexity sequences will be associated with larger gradients, making the sequence easier to remember~\cite{carlini2022privacy, feldman2020does}.

We consider two strategies to generate trap sequences: using $LM_{\text{ref}}$ to generate synthetic sequences ($M_{D, \text{synth}}$) and sampling existing sequences from the document $D$ ($M_{D, \text{real}}$).

\begin{figure}
\centering
\includegraphics[width=0.5\linewidth]{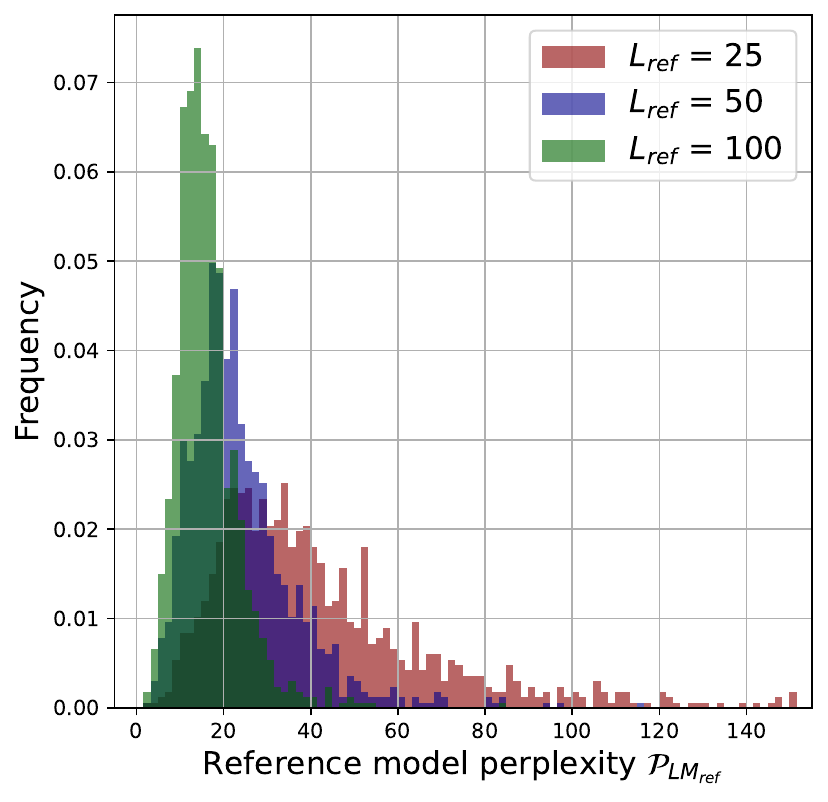} 
    \caption{The distribution of reference model perplexity $\mathcal{P}_{\textit{LM}_{\text{ref}}}$ computed on 1,000 sequences each of length $L_{\text{ref}}(M_D) = \{25, 50, 100\}$. The sequences are randomly sampled from the 500 books in $D_\textit{NM}$ (see Sec.~\ref{sec:dataset})} 
\label{fig:seq_length_ppl}
\end{figure}

For $M_{D, \text{synth}}$, we start with an empty prompt and use $LM_{\text{ref}}$ to generate tokens using top-$k$ sampling ($k=50$) until reaching the target length. For increased diversity of samples we vary the \textit{temperature} $t = \{0.5, 1.0, \dots, 8.0\}$. For $M_{D, \text{real}}$, we sample sequences of a given length directly from the document $D$. We repeat the process until we have 50 trap sequences per bucket $b_i$ for $i = 1 \dots 10$, with any excess sequences discarded. We provide examples of synthetically generated trap sequences in Appendix~\ref{app:examples}. To illustrate the perplexity range we here consider, Fig.~\ref{fig:seq_length_ppl} shows the perplexity distribution of randomly sampled sequences from real books.

\subsection{Dataset of books in the public domain}
\label{sec:dataset}

We inject the trap sequences at random in a homogeneous dataset of text. More specifically, we use the open-source library~\cite{kpullygutenberg} to collect 9,542 books made available in the public domain on Project Gutenberg~\cite{projectgutenberg} which were not included in the PG-19 dataset~\cite{raecompressive2019}. We only consider books with at least 5000 tokens using the tokenizer from reference model LLaMA-2 7B~\cite{touvron2023llama2}. The length of the selected books follows a heavy tail distribution, with a mean of 98k and 90-percentile of 204k tokens. Note that these books have no overlap with the rest of the training dataset.

To ensure a controlled setup for document-level membership inference, we consider two random subsets of books from this collection in which no trap sequences are injected. We designate one part as \emph{non-members}, $D_\textit{NM}$ of size $|D_\textit{NM}|=500$, excluded from the training dataset and \emph{members}, $D_\textit{M}$ of size $|D_\textit{M}|=500$, which are included in the training dataset in its original form, i.e. $D=D'$. 

\subsection{Trap sequence injection}
\label{sec:m_injection}

To inject trap sequence $M_D$ into a book $D$, we first split the book by spaces, ensuring injections are not splitting existing words. We then select $n_{\text{rep}}$ random splits, in each of which $M_D$ is injected, resulting in modified document $D'$.

We create modified books $D'$ using $M_{D, \text{synth}}$ and $M_{D, \text{real}}$ as described in Sec.~\ref{sec:m_generation}. On top of varying the sequence length and perplexity bucket for each $M_{D}$, we also vary the number of times it is injected into document D: $n_{\text{rep}} = \left\{1, 10, 100, 1000\right\}$ for $M_{D, \text{synth}}$ and $n_{\text{rep}} = 100$ for $M_{D, \text{real}}$. We consider 50 sequences per combination of $(L_{\text{ref}}, b_i, n_{\text{rep}})$ and only inject one unique $M_{D}$ per book, resulting in a set of $7,500$ randomly picked books each containing trap sequences.

\subsection{Training of the target LLM}
\label{sec:model_training}
The LLM we target in this project is part of a larger effort to train a highly efficient model of relatively small size (1.3B parameters), on a large training set consisting of 3 trillion tokens of English, French and Code data~\cite{faysse2024croissantllm}. In line with recent work~\cite{touvron2023llama}, this model is trained to be ``inference-optimal''. This means that compute allocation and model design decisions were driven by the objective of having the best model possible for a given number of parameters, rather than the best possible model for a given compute budget~\cite{hoffmann2022training}.  

We here provide a high-level overview of the LLM training characteristics, but refer to the technical report for more details~\cite{faysse2024croissantllm}. 

\textbf{Data.} The training corpus consists of content associated with free-use licences, originating from filtered internet content, as well as public domain books, encyclopedias, speech transcripts and beyond. Data is upsampled at most twice for English data, which has been shown to lead to negligible performance decrease with respect to non-upsampled training sets~\cite{muennighoff2023scaling}. The final dataset represents 4.1 TB of unique data.

\textbf{Copyright trap inclusion.} Trap sequences are disseminated within the model training set and seen twice during training. In total, documents containing trap sequences represent less than 0.04 \% of tokens seen by the model during training, minimizing the potential impact of including trap sequences on our model performance.

\textbf{Tokenizer.} The tokenizer is a BPE SentencePiece tokenizer fitted on a corpus consisting of 100 billion tokens of English, French and Code data. It has a vocabulary of 32,000 tokens with white space separation and byte fallback.

\textbf{Model.} The model is a 1.3 billion parameter LLaMA model~\cite{touvron2023llama2} with 24 layers, a hidden size of 2,048, an intermediate size of 5,504 and 16 key-value heads. It is trained with Microsoft DeepSpeed on a distributed compute cluster, with 30 nodes of 8 x Nvidia A100 GPUs during 17 days. Training is done with a batches of 7,680 sequences of length 2,048, which means that over 15 million tokens are seen at each training step.

\textbf{Model Performance.} Evaluation of the final models suggest very strong performance for its size, edging out similarly sized models~\cite{biderman2023pythia,zhang2022opt,scao2022bloom} on English benchmarks and largely surpassing them on French benchmarks. 

\subsection{Setup for trap sequence MIA}
\label{sec:mia_setup}

In order to infer whether document $D'$ containing trap sequence $M_D$ has been used to train target model $LM$, we implement sequence-level Membership Inference Attacks (MIAs) from the literature. 

As \emph{members}, we consider the trap sequences, both $M_{D, \text{synth}}$ and $M_{D, \text{real}}$, which we created and injected as described in Sec.~\ref{sec:m_generation} and Sec.~\ref{sec:m_injection} - as they all have been included in the training dataset of $LM$. 

As \emph{non-members}, we repeat the exact same generation process to create a similar set of sequences that we exclude from the training dataset. For $M_{D, \text{synth}}$, this means repeating the same top-$k$ sampling approach with a different random seed, until the same number of sequences is collected for each combination $(L_{\text{ref}}, b_i)$. For $M_{D, \text{real}}$, we use randomly sampled sequences from $D_\textit{NM}$ as described in Sec.~\ref{sec:dataset}.

We consider $X$ as any sequence, which is either \emph{member} or \emph{non-member}, and aim to infer whether $X \in \mathcal{D}_{\text{train}}$ or not. We select three methods for sequence-level MIA to compute an \emph{attack score} $\alpha$: 

\textbf{\textit{1. Loss}} attack from~\cite{yeom2018privacy}, which uses the model loss $\alpha = \mathcal{L}_{\textit{LM}}(X)$. 

\textbf{\textit{2. Min-K\% Prob}} from~\cite{shi2023detecting}, which computes the mean log-likelihood of the k\% tokens with minimum predicted probability in the sequence. More formally, $\alpha=\frac{1}{E}\sum_{t_i \in Min-K\%} \log\left( \textit{LM}_{\theta}(t_i)\right)$, where $E$ is the number of tokens in $Min-K\%$ and we consider $k=20$.

\textbf{\textit{3. Ratio}} attack from~\cite{carlini2021extracting}, which uses the model loss divided by the loss computed using a reference model, or $\alpha = \mathcal{L}_{\textit{LM}}(X) / \mathcal{L}_{\textit{LM}_{\text{ref}}}(X)$. We use the same $\textit{LM}_{\text{ref}}$ as used to generate synthetic trap sequences, i.e. LLaMA-2 7B~\cite{touvron2023llama2}.

We compute the attack score $\alpha$ for a balanced membership dataset of trap sequences and similarly generated non-member sequences, which is then used to calculate the AUC of the binary membership prediction task.

Importantly, the setup described above allows us, unlike prior work~\cite{carlini2022quantifying,kandpal2022deduplicating}, to draw causal conclusions about memorization and factors affecting it. Where "natural experiments" could suffer from known or unknown confounding factors, we here generate (Sec.~\ref{sec:m_generation}) and inject (Sec.~\ref{sec:m_injection}) trap sequences randomly, thus guaranteeing any observed difference in loss is explained solely by a controlled injection into the training dataset and subsequent memorization. This enables us to draw causal conclusions between perplexity and memorization, while we find post-hoc analyses to likely be impacted by perplexity as a confounding factor (Sec.~\ref{sec:perplexity}). 

\section{Results}

\subsection{Recent document-level MIAs are not sufficient}

We first only consider books in with no trap sequences injected, for which we have non-member $D_\textit{NM}$ and member $D_\textit{M}$ documents as stated in Sec.~\ref{sec:dataset}. This allows us to implement two methods proposed in prior work to infer document-level membership for LLMs. 

First, we implement the method from~\cite{meeus2023did}. We query \textit{LM} with context length $C=1024$, and use as normalization strategy $\textit{MaxNormTF}$ and as feature extractor a histogram with $500$ bins. We split the dataset of books in $h = 5$ chunks, each consisting of a random subset of $100$ member and non-members, and train $h$ meta-classifiers on $h - 1$  chunks to be evaluated on the held out chunk. 

Second, we implement the \textit{Min-K\% Prob} from~\cite{shi2023detecting}. Following the proposed setup for books, we sample 100 random excerpts of 512 tokens from each book and compute the Min-K\% Prob for each excerpt with $k=20$. The sequence-level threshold for binary prediction is determined to maximize accuracy. The average prediction per book then serves as predicted probability for membership, and used to compute an AUC. We repeat this process $h=5$ times, sampling excerpts with a different random seed. 

Table~\ref{tab:document_level_MIA} summarizes the results. Notably, the AUC for both methods is barely above the random guess baseline, while in their original setup the methods achieved an AUC of 0.86~\cite{meeus2023did} and 0.88~\cite{shi2023detecting}. This confirms our hypothesis that the LLM we here consider is significantly less prone to memorization than the models used in prior work. Our 1.3B model has been trained on 4TB on data, while for instance LLaMA 7B -a representative target model for both methods- contains 6 times as many parameters while trained on a dataset of a similar size (4.75TB)~\cite{touvron2023llama}. In line with the trends confirmed in prior work~\cite{carlini2022quantifying,shi2023detecting}, having less parameters and a large dataset size suggest our model to be less prone to memorization. 

These results show that for many training setups, LLMs do not exhibit memorization to the extent necessary to make state-of-the-art methods in document-level membership inference succeed. They are thus not sufficient to help content creators verify the use of their documents to train LLMs, emphasizing the need for novel approaches such as ours. 

\begin{table}[t]
    \centering
    \begin{tabular}{ccc}
    \toprule
        Method & AUC \\
        \midrule 
        \cite{meeus2023did} & $0.513 \pm 0.021$ \\ 
        \cite{shi2023detecting} & $0.524 \pm 0.003 $ \\
        \bottomrule
    \end{tabular}
    \caption{Mean and standard deviation of an AUC for the document-level inference on books not containing any trap sequences.}
    \label{tab:document_level_MIA}
\end{table}

\subsection{Sequence-level MIA for synthetically generated trap sequences}

We approach the task of document-level membership inference with a sequence-level MIA, with injected trap sequences as members and similarly generated sequences as non-members as described in Sec.~\ref{sec:mia_setup}. Table~\ref{tab:synt_roc_auc} summarizes the AUC for all MIA methodologies considered, when applied to the synthetically generated trap sequences $M_{D, \text{synth}}$ across sequence length $L_\text{ref}$ and number of repetitions $n_\text{rep}$. 

Contrary to popular intuition~\cite{carlini2022quantifying, kandpal2022deduplicating}, we show that repeating a sequence large number of times does not easily lead to memorization. Indeed, even for a reasonably long sequence of 50 tokens, 100 duplicates is not enough to make it reliably detectable by any of the methods we consider. For $L_\text{ref}=25$, even $1,000$ repetitions is not sufficient. 

We had hypothesized that detectability might be affected by fact that trap sequences bear no semantic connection to the document $D$. This could potentially lead to the sequence being an extreme outlier and virtually discarded during the training process, as LLMs are typically trained on the noisy data and generally robust to outliers. 

To test this hypothesis, we therefore sampled trap sequences $M_{D, \text{real}}$ from the same distribution as the document $D$ and injected them in our training set. In practice, this means repeating an excerpt from $D$ $n_\text{rep}$ number of times. However, we find that, similarly to synthetically generated sequences, $L_\text{ref}=50$ tokens repeated $n_\text{rep}=100$ times is not sufficient to make the MIAs perform reliably better than chance, with the resulting AUC of the \textit{Ratio} MIA of 0.492. This disproves the outlier hypothesis and confirms that detectability is harder than one might think.

Increasing the sequence length and/or number of repetitions however allows the trap to be memorized and, consequently, detected with an AUC of up to 0.748 for sequence length $L_\text{ref} = 100$, repeated $n_\text{rep} = 1,000$ times. Decreasing the number of repetitions to $n_\text{rep} = 100$ ($L_\text{ref} = 100$) decreases the AUC to 0.639 while $L_\text{ref} = 50$ ($n_\text{rep} = 1,000$) decreases it to 0.627. 

Excitingly, these results show that trap sequences can enable content detectability even in models that would not naturally memorize including small models such as the ones used on device, giving creators an opportunity to verify whether their content was seen by a model. To be effective, however, current trap sequences need to be long and/or repeated a large number of times. The inclusion on sequence traps therefore relies (see Sec.~\ref{sec:discussion}) on the ability of the content creator to include them in the content in a way that does not impact its readability e.g. text that would be collected by a scraper but not visible to users. 

\newcolumntype{C}[1]{>{\centering\arraybackslash}m{#1}}

\begin{table}[ht]
    \centering
        \begin{tabular}{C{0.7cm} C{0.7cm} C{1.3cm} C{1.3cm} C{1.3cm}}
    \toprule
         $L_\text{ref}$ & $n_\text{rep}$ & \textit{Loss} & \textit{Min-K\% Prob} & \textit{Ratio} \\
         \midrule
         \multirow{5}{*}{25} & 1 & $0.454$ & $0.461$ & $0.490$ \\ 
         \cmidrule{2-5}
         & 10 & $0.508$ & $0.515$ & $0.515$ \\ 
         \cmidrule{2-5}
         & 100 & $0.520$ & $0.545$ & $0.524$ \\ 
         \cmidrule{2-5}
         & 1000 & $0.548$ & $0.539$ & $0.557$ \\ 
         \midrule
         \multirow{5}{*}{50} & 1 & $0.462$ & $0.496$ & $0.510$ \\ 
         \cmidrule{2-5}
         & 10 & $0.505$ & $0.543$ & $0.506$ \\ 
         \cmidrule{2-5}
         & 100 & $0.520$ & $0.515$ & $0.521$ \\ 
         \cmidrule{2-5}
         & 1000 & $0.562$ & $0.610$ & \bm{$0.627$} \\ 
         \midrule
         \multirow{5}{*}{100} & 1 & $0.482$ & $0.463$ & $0.550$ \\ 
         \cmidrule{2-5}
         & 10 & $0.529$ & $0.502$ & $0.552$ \\ 
         \cmidrule{2-5}
         & 100 & $0.562$ & $0.546$ & \bm{$0.639$} \\ 
         \cmidrule{2-5}
         & 1000 & $0.611$ & $0.599$ & \bm{$0.748$} \\ 
         \bottomrule
    \end{tabular}
    \caption{MIA AUC for synthetic trap sequences. Each AUC value is computed using 500 members and 500 non-members, equally distributed across reference model perplexity buckets $b_i$. }
\label{tab:synt_roc_auc}
\end{table}

\subsection{MIA performance during model training}

As described in Sec.~\ref{sec:model_training}, we train the 1.3B target model from scratch. As part of the training, we also save intermediate model checkpoints every 5,000 training steps (for each step 15M tokens are seen by the model). As the dataset is shuffled before training, the trap sequence occurrences are uniformly distributed within the epoch, allowing us to perform a post-hoc study on the memorization throughout the training process. We perform the sequence-level MIAs on a series of model checkpoints and report the AUC. 

Figure~\ref{fig:across_training} contains the MIA results across training for synthetically generated trap sequences $M_{D, \text{synth}}$ for varying sequence lengths $L_\text{ref}$. We here consider \textit{Ratio} attack and $n_\text{rep}=1,000$.

Notably, the AUC increases smoothly and monotonically for model checkpoints further in the training process. This demonstrates the relationship between the detectability and a number of times the model has seen a trap sequence, which increases linearly with training steps. We also observe that the AUC has not yet reached a plateau and would likely further increase if more training steps were included. We hypothesize that LLM developers could also measure -and extrapolate- LLM detectability over training through MIAs on injected sequences, which we leave for future work to explore. These results shed light in how the detectability of specific sequences evolves for a real-world LLM, which -to our knowledge- is not documented by prior work. 

\subsection{Perplexity and detectability}
\label{sec:perplexity}

As a part of our experiment design (Sec.~\ref{sec:m_generation}), we investigate a hypothesis that, in addition to the length and the number of repetitions, detectability of a trap sequence depends on its perplexity (computed by a reference model). 

\begin{figure}[ht]
\centering
\includegraphics[width=0.5\linewidth]{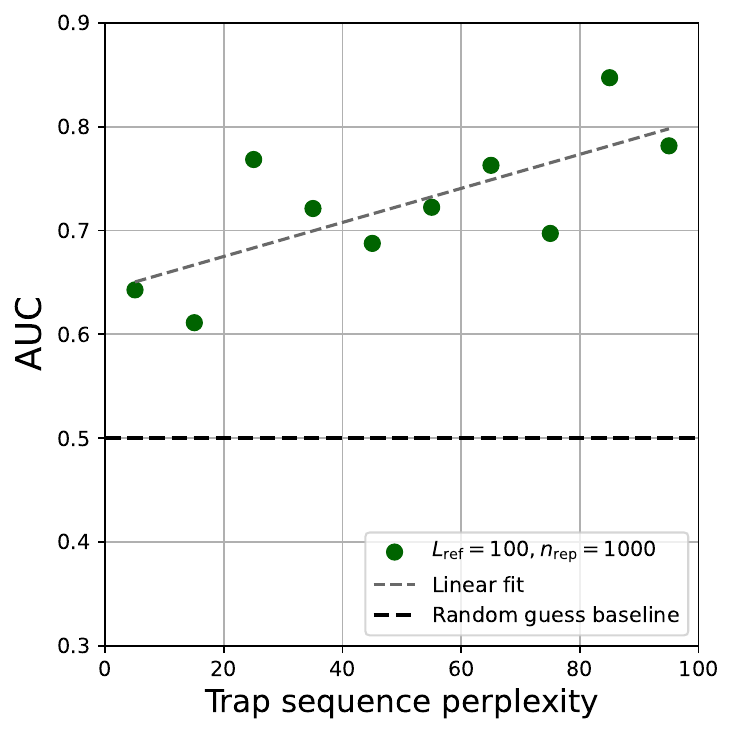} 
    \caption{The relationship between \textit{Ratio} MIA AUC and trap sequence perplexity (bucketized) in the $L_\text{ref}=100$, $n_\text{rep}=1000$ setup. Pearson correlation coefficient is 0.715 with p-value = 0.02.}
\label{fig:ppl_correlation}
\end{figure}

We focus on the setup with the highest level of memorization observed: $L_\text{ref}=100, n_\text{rep}=1,000$ and consider the AUC reported by the best performing MIA (Figure~\ref{fig:ppl_correlation}). Indeed, we find a positive correlation between the AUC and the trap sequence perplexity (bucketized as per Sec.~\ref{sec:m_generation}) with a Pearson correlation coefficient of 0.715 and significant p-value (0.02). Compared to naturally occurring sequences of $L_\text{ref} = 100$ (Fig.~\ref{fig:seq_length_ppl}), the most detectable sequences have much higher perplexity. These results allow us to conclude that, in general, 'outlier' sequences tend to be more detectable after training, even if the perplexity is determined by an unrelated reference model. 

\begin{figure}[ht]
\centering
\includegraphics[width=0.5\linewidth]{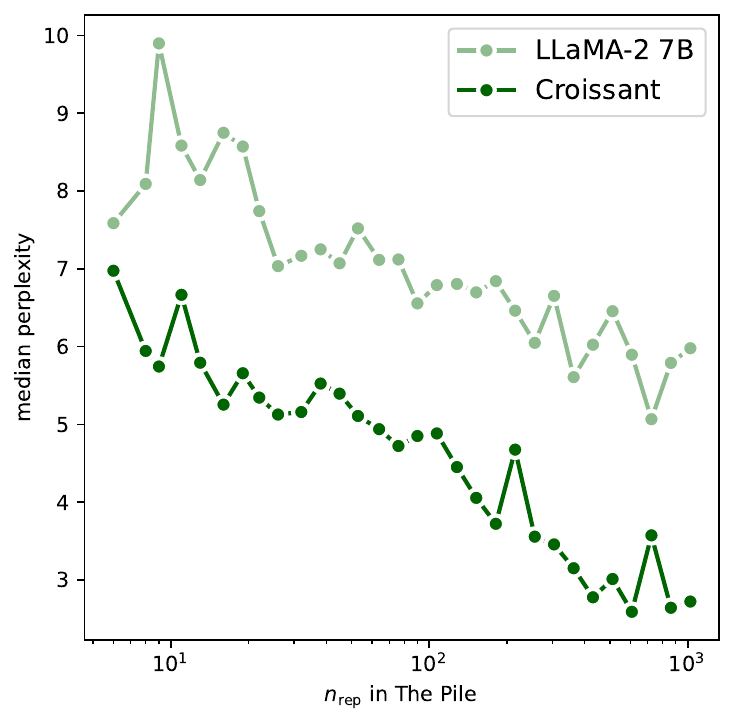} 
    \caption{Perplexity of naturally occurring duplicates in The Pile. Each duplicate is a sequence of 100 GPT-2 tokens, repeated $n_\text{rep}$ times in the dataset. Each data point represents a median of 100 randomly drawn samples.}
\label{fig:natural_duplicates_ppl}
\end{figure}

To put this in the context of prior work, we compute the perplexity of naturally occurring duplicates in The Pile~\cite{pile}, previously used to quantify LLM memorization~\cite{carlini2022quantifying}. We use the code provided by~\cite{lee2022deduplicating} to identify sequences of 100 (GPT-2) tokens repeated between $6$ to $1,024$ times in the non-copyrighted version of The Pile - where all of the copyright-protected content comprising roughly 20\% of the original datset was removed~\cite{pile_uncopyrighted}. We then compute the perplexity of such sequences with LLaMA-2 7B (our reference model), and CroissantLLM (the model we here train). Fig.~\ref{fig:natural_duplicates_ppl} shows that sequences repeated more often also tend to have lower perplexity. Thus, according to our findings above, they are also easier for the model to memorize - making perplexity a potential and unexplored confounding factor in post-hoc analyses. It is important to note, however, that the observed decrease in perplexity with repetition presented in Fig.~\ref{fig:natural_duplicates_ppl} could also be partially attributed to memorization. While neither of the models has been explicitly trained on The Pile, it is possible that frequently repeated sequences in The Pile also tend to be prevalent across other large text datasets, potentially leading to memorization (lower perplexity) by both models. We therefore argue that these results highlight the challenges in studying memorization post-hoc, and underscore the importance of randomized controlled studies, such as presented in this paper.

\subsection{Leveraging the context}

Performing a sequence-level MIA for trap sequences as a proxy for document-level MIA does not fully leverage the knowledge available to an attacker, i.e. the context in which the sequences appear. We here evaluate whether we can improve the MIA performance when we compute the model loss when also providing the corresponding context. 
First, for each trap sequence $M_D$, we randomly sample one occurrence of $M_D$ in $D'$ (out of $n_\text{rep}$). From this location in the original document $D$, we retrieve the textual context $C$ of length $L_\text{ref}(C)$ tokens preceding the injected sequence. We can then compute the model loss of the sequence $X$ in this context, 
$
\mathcal{L}_{\textit{LM}}(X, C) = -\frac{1}{L}\sum_{i=1}^{L} \log\left( \textit{LM}_{\theta}(t_i | T(C), t_1 \ldots, t_{i-1})\right)$ where $T(C)$ corresponds to the tokenized context. Considering sequence $X$, which is either the injected $M_D$ or a similarly created sequence not injected, we use the modified \textit{Ratio} attack with $\alpha = \mathcal{L}_{\textit{LM}}(X, C) / \mathcal{L}_{\textit{LM}_{\text{ref}}}(X, C)$. 

Table~\ref{tab:context} shows how the MIA AUC changes when we consider a context of $L_\text{ref}(C)=100$ tokens. We find that for short and medium-length sequences, the MIA performance tends to increase when context is taken into account, while for longer sequences, it remains fairly similar. These results suggest that more effective ways of leveraging the context may exist, effectively bridging the gap between MIAs applied on the trap sequence and document-level. Lastly, these results also suggest that the context in which naturally occurring duplicates occur could be another confounding factor in post-hoc memorization studies. We leave this to future work to explore. 

\begin{table}[ht]
    \centering
        \begin{tabular}{C{0.5cm} C{0.5cm} C{2.1cm} C{2.1cm}}
    \toprule
         $L_\text{ref}$ & $n_\text{rep}$ & No context & $L_\text{ref}(C)=100$ \\
         \midrule
         \multirow{4}{*}{25} & 10 & $0.515$ & \bm{$0.534$} \\ 
         \cmidrule{2-4}
         & 100 & $0.524$ & \bm{$0.535$} \\ 
         \cmidrule{2-4}
         & 1000 & $0.557$ & \bm{$0.603$} \\ 
         \midrule
         \multirow{4}{*}{50} & 10 & \bm{$0.506$} & $0.500$ \\ 
         \cmidrule{2-4}
         & 100 & $0.521$ & \bm{$0.581$}\\ 
         \cmidrule{2-4}
         & 1000 & $0.627$ & \bm{$0.685$} \\ 
         \midrule
         \multirow{4}{*}{100} & 10 & \bm{$0.552$} & $0.531$ \\ 
         \cmidrule{2-4}
         & 100 & $0.639$ & \bm{$0.642$} \\ 
         \cmidrule{2-4}
         & 1000 & \bm{$0.748$} & $0.739$ \\ 
         \bottomrule
    \end{tabular}
    \caption{\textit{Ratio} MIA AUC for synthetic trap sequences, comparing the results without context and considering a context of 100 tokens.}
\label{tab:context}
\end{table}

\subsection{Impact of parameter precision}

We now study how potential memorization mitigation strategies would impact trap detectability. Specifically, we perform our best available MIA (\textit{Ratio}) for $L_\text{ref}=100$ and $n_\text{rep}=1,000$, on the target model $\textit{LM}$ loaded with different precision of model parameters. Thus far, we only considered a floating point precision of 32 bits (\textit{float32}), and we now additionally include floating point precision of 16bits (\textit{float16}) and integer precision of 8 and 4 bits (\textit{int8}, \textit{int4}). 

Tab.~\ref{tab:precision} shows how the MIA AUC changes with the target model parameter precision. Unsurprisingly, as we hypothesize parameter precision to be related with model's capacity to memorize, we find that the AUC decreases slowly for decreasing precision. However, even when the model is loaded with integer precision of 4 bits, we find the AUC of $0.70$ to be significantly above the random guess baseline, suggesting that copyright traps remain effective even when memorization mitigation strategies are employed. 

\begin{table}[ht]
    \centering
        \begin{tabular}{C{2cm} C{2cm}}
    \toprule
         $\textit{LM}$ precision & AUC \\
         \midrule
         \textit{float32} &  $0.748$ \\ 
         \textit{float16} &  $0.745$ \\ 
         \textit{int8} &  $0.738$ \\ 
         \textit{int4} &  $0.697$ \\ 
         \bottomrule
    \end{tabular}
    \caption{\textit{Ratio} MIA AUC for synthetic trap sequences with $L_\text{ref}=100$ and $n_\text{rep}=1,000$, across model's parameter precision.}
\label{tab:precision}
\end{table}


\section{Discussion and Future Work}

\textbf{Data preprocessing.} Clean and high-quality training data is increasingly recognized as a key component in training LLMs~\cite{lee2022deduplicating}. One of the most commonly deployed preprocessing steps is data deduplication. Our proposed method relies on repeating trap sequences many times, and is therefore sensitive to a sequence level deduplication. We, however, believe the method to be relevant now, and in the foreseeable future. Most deduplication is performed on a document-level~\cite{cerebras2023slimpajama,penedo2023refinedweb}, which does not interfere with our method. Sequence-level deduplication has also been proposed, but suffers from fundamental drawbacks. First, it is very computationally expensive, especially for large datasets containing terabytes of text~\cite{lee2022deduplicating}. Prior work has also shown deduplication to have negative impact on performance on certain tasks~\cite{roberts2020much}, making aggressive deduplication potentially detrimental for model utility. Further, developers have employed rule-based~\cite{kudugunta2024madlad,scao2022bloom} and perplexity~\cite{wenzek2019ccnet} filters, both of which we find not to affect injected trap sequences. 



\textbf{Readability.} Apart from detectability, content readability is an important practical implication of copyright traps. In our experiments we show that only injecting a relatively long sequence up to a 1,000 times leads to significant impact on detectability. While this may not be practical for some content creators (e.g. book authors), we believe this is feasible for some creators in its current form. For instance online publishers could include sequences across articles, invisible to the users, yet appearing as rendered text to a web-scraper. As a proof of concept, we have incorporated a trap into an invisible HTML element and confirmed that it was successfully retrieved by an Apache Nutch web crawler - also used for Common Crawl~\cite{commoncrawl}.
Beyond that, this work presents early research towards document-level inference, and we expect more progress towards the practical solution in the future. 

\textbf{Relation to backdoor attacks.} Backdoor attacks rely on a hidden trigger embedded in the training data of machine learning models, typically with the aim of inducing a desired (mis)classification of data containing similar triggers at inference time. 
In contrast, the copyright traps we here propose do not aim to trigger specific classifications in the target model's output and are designed to enhance detectability in LLM training data. Future work could explore how techniques proven to be successful as backdoor attacks could be used for similar purposes. 
\label{sec:discussion}

\section{Conclusion}
With the copyright concerns regarding LLM training being raised, LLM developers are reluctant to disclose details on their training data. Prior work has explored the question of document-level membership inference to detect whether a piece of content has been used to train a LLM. We first show that memorization highly depends on the training setup, as existing document-level membership inference methods fail for our 1.3B LLM. We thus propose the use of copyright traps for LLMs - purposefully designed text sequences injected into a document, intended to maximize detectability in LLM training data. 

We train a real-world, 1.3B LLM from scratch on 3 trillion tokens, containing a small set of injected trap sequences, enabling us to study their effectiveness. We find that inducing reliable memorization in a LLM is a non-trivial task. For models showing relatively low level of memorization, such as the one we train here, injecting short-to-medium sentences ($\leq $ 50 tokens) up to a 100 times does not improve document detectability. When using longer sequences, however, and up to a $1,000$ repetitions, we do see a significant effect - showing how copyright traps can enable detactability even for LLMs less prone to memorize. We further find that memorization increases with sequence perplexity, and that leveraging document-level information such as context could boost detectability. While effective, the proposed mechanism could be disruptive to the document's content and readability. Future research is thus needed, specifically in designing trap sequences maximizing detectability. We are hence committed to releasing our model and the data to further the research in the field.



\section*{Availability}
The target LLM, CroissantLLM, is readily accessible on HuggingFace\footnote{\url{https://huggingface.co/croissantllm}}. The entire training dataset for CroissantLLM will be made publicly available too, including the trap sequences. The code used for trap sequence generation and analysis is available on github\footnote{\url{https://github.com/computationalprivacy/copyright-traps}}.


\section*{Impact Statement}
While the exact legal nature of copyright in the context of LLM training is still actively debated, the study of copyright traps increases transparency in model training. We believe this to be generally beneficial for the community of content creators, researchers and model developers. 

It is worth noting, however, that openly publishing this research would make it easier for malevolent model developers to evade any potential measures to increase the detectability of the training data, should they be developed.

More broadly, this work also contributes to the large body of research dedicated at exploring training data extraction, which can be a serious privacy threat. By exploring which properties affect memorization in a real-world LLM, we believe to effectively contribute to understanding the associated privacy risk - which is beneficial for both model developers aiming to limit privacy threats and society as a whole. We believe that further exploration of the topic does not pose additional risks, as privacy risks in LLMs mostly come from unintended memorization, rather than a deliberate malice by a model developer.

On the other hand, we find that memorization capacity varies greatly across different models, and not all models are equally prone to memorizing their training data. We hope that this finding does not lead to an increased complacency to privacy concerns among model developers.

Separately, potential misuse of LLMs for producing misinformation should be considered. Better understanding of LLM memorization could be abused by bad actors to influence the output of production-grade LLMs. We, however, believe that the benefits of the research in this area outweigh the risks, and it will help inform future defences against misuse. 


\section*{Acknowledgements}
Training compute is obtained on the Jean Zay supercomputer operated by Genci Idris through compute grant 2023-AD011014668R1.

\bibliography{bibliography}
\bibliographystyle{icml2024}

\newpage
\appendix
\onecolumn

\section{Appendix: Example Trap Sequences}
\label{app:examples}

Table~\ref{tab:examples} shows examples of synthetically generated trap sequences $M_D$ for varying length $L_{\text{ref}}(M_D)$ and perplexity $\mathcal{P}_{\textit{LM}_{\text{ref}}}(M_D)$ computed using reference language model $\textit{LM}_{\text{ref}}$. 

\begin{table*}[ht]
    \centering
    \begin{tabular}{C{1.5cm}C{3cm}p{10cm}}
    \toprule
        $L_{\text{ref}}(M_D)$ & $\mathcal{P}_{\textit{LM}_{\text{ref}}}(M_D)$ & \multicolumn{1}{c}{Trap sequence $M_D$} \\
        \midrule
        25 & $1 \leq \mathcal{P}_{\textit{LM}_{\text{ref}}} < 11$ & \textit{"It’s my favorite time of the year: the time between New Year’s and Easter; there are so many"}\\
        \midrule
        25 & $41 \leq \mathcal{P}_{\textit{LM}_{\text{ref}}} < 51$ & \textit{"When it comes for designing inter-connected solutions in different disciplines (meeting room solution, virtual workplace, conference"}\\
        \midrule
        25 & $91 \leq \mathcal{P}_{\textit{LM}_{\text{ref}}} < 101$ & \textit{"If you are the proprietary, you want an app in your organization store. On April four,, all of those individuals affected"}\\
        \midrule
        50 & $1 \leq \mathcal{P}_{\textit{LM}_{\text{ref}}} < 11$ & \textit{"If you or someone you know has been charged with a crime in West Palm Beach, you'll want an experienced criminal defense attorney. If you don't have an attorney, the first thing you should do is to talk to a"}\\
        \midrule
        50 & $41 \leq \mathcal{P}_{\textit{LM}_{\text{ref}}} < 51$ & \textit{"As we go from one 9:30am kick off into two over here, then, we’ve got you more-ish for every game and then even if all our football was getting in your backbone after a night away, we"}\\
        \midrule
        50 & $91 \leq \mathcal{P}_{\textit{LM}_{\text{ref}}} < 101$ & \textit{"When in comes times of turmoil... whats on sale and more important when, is best, this list tells your who is opening on Thrs. at night with their regular sale times and other opening time from your neighbors. You still"}\\
        \midrule
        100 & $1 \leq \mathcal{P}_{\textit{LM}_{\text{ref}}} < 11$ & \textit{"A few days ago, I started a new exercise routine. It's been a few years since I've been serious about working out. I'm doing this to get in shape for a trip to Italy in the spring.
        Today, I went to the gym for the first time. I didn't feel any pain, and I did everything I was supposed to do.
        I felt really good afterward. I wasn't sure what to expect. I thought"}\\
        \midrule
        100 & $41 \leq \mathcal{P}_{\textit{LM}_{\text{ref}}} < 51$ & \textit{".You don't care? But it has to be a big enough carpet square as a base so we can easily hide the paw marks. 0: No! They don' t need a base at
        You don't want your puppy at all? If I bring you some other time
        Maybe, we bring another dog? Then maybe this puppy and maybe that
        dog (purring, lick) we just play fetch for
        a couple"}\\
        \midrule
        100 & $91 \leq \mathcal{P}_{\textit{LM}_{\text{ref}}} < 101$ & \textit{"Founded over seventieth FAR FAR and then over the year Fashion-Ralph became so established on online, fashion.net became on the mainland for us that there are few competitors we had as a name in Italy , then after several online stores of the best online for each item (such as ToskanaT-Tops Italy for fashion , La Marcia of accessory ), then all became united on a market network with one store but several pages ."}\\
        \bottomrule
    \end{tabular}
    \caption{Example of synthetically generated trap sequences for varying length $L_{\text{ref}}(M_D)$ and perplexity $\mathcal{P}_{\textit{LM}_{\text{ref}}}(M_D)$.}
    \label{tab:examples}
\end{table*} 


\end{document}